%% file: main.tex
\documentclass[10pt,letterpaper]{article}
\usepackage{arxiv}
\usepackage[utf8]{inputenc}
\usepackage[dvipsnames]{xcolor}
\usepackage[numbers,sort&compress]{natbib}
\usepackage[shortlabels,inline]{enumitem}
\usepackage[font=small,labelfont=bf]{caption}
\usepackage[hyphens]{url}
\input{preamble}

\definecolor{linkblue}{rgb}{0.21,0.49,0.74}
\usepackage[breaklinks,colorlinks,allcolors=linkblue]{hyperref}
\hypersetup{pdftitle={EviBridge: Encoding Predictive Visual Evidence for Proactive Interaction},pdfauthor={Zhichao Huang, Changheng Cai, Song Ren, Zhichen Zhao, Hongsen Liu, Chaoyang Zheng, Wenxi Wu, Yichen Long, Yuan Fang, Luya Wang, Gufeng Liu, Yuchen Wang}}
\renewcommand{\shorttitle}{\eb{}}
\renewcommand{\headeright}{Preprint}
\renewcommand{\undertitle}{Preprint}
\title{EviBridge: Encoding Predictive Visual Evidence for Proactive Interaction}
\author{\parbox{\dimexpr\textwidth-2\tabcolsep\relax}{\centering\normalfont
Zhichao Huang\textsuperscript{1}, Changheng Cai\textsuperscript{1}, Song Ren\textsuperscript{1}, Zhichen Zhao\textsuperscript{2},\\[3pt]
Hongsen Liu\textsuperscript{1}, Chaoyang Zheng\textsuperscript{1}, Wenxi Wu\textsuperscript{1}, Yichen Long\textsuperscript{1},\\[3pt]
Yuan Fang\textsuperscript{1}, Luya Wang\textsuperscript{1}, Gufeng Liu\textsuperscript{1}, Yuchen Wang\textsuperscript{1}\\[8pt]
\small\textsuperscript{1}JD.com\\[2pt]
\textsuperscript{2}The Hong Kong University of Science and Technology
}}
\date{}
\begin{document}
\maketitle
\suppressfloats[t]
\input{sec/0_abstract}
\input{sec/1_intro}
\input{sec/2_related}
\input{sec/3_method}

\input{sec/4_experiments}
\input{sec/5_conclusion}
\section*{LLM Usage Statement}
We used Codex to help write parts of the code and polish parts of the manuscript.
\bibliographystyle{plainnat}
\begingroup
\raggedright
\bibliography{main}
\endgroup
\end{document}

%% file: preamble.tex
\usepackage[T1]{fontenc}

\newcommand{\eb}{\textup{\textsc{EviBridge}}}
\newcommand{\fgi}{\textup{\textsc{FGI}}}
\newcommand{\bench}{\textup{\textsc{FuGIBench}}}
\newcommand{\wait}{\textup{\textsc{Wait}}}
\newcommand{\prepare}{\textup{\textsc{Prepare}}}
\newcommand{\response}{\textup{\textsc{Response}}}
\newcommand{\waittok}{\texttt{</fgi\_wait>}}
\newcommand{\preparetok}{\texttt{</fgi\_prepare>}}
\newcommand{\responsetok}{\texttt{</fgi\_response>}}

\usepackage{amsmath,amssymb}
\usepackage{booktabs}
\usepackage{colortbl}
\definecolor{evirow}{RGB}{226,238,248}
\usepackage{graphicx}
\usepackage{multirow}
\usepackage{tikz}
\usetikzlibrary{arrows.meta,positioning,fit,calc}
\usepackage{xspace}
\usepackage{microtype}
\usepackage{enumitem}

\newcommand{\ebe}{\textup{\textsc{EviBridge-E}}\xspace}
\newcommand{\ebi}{\textup{\textsc{EviBridge-I}}\xspace}

\newcommand{\best}[1]{\textbf{#1}}
\newcommand{\bestcap}[1]{\textbf{#1}}

%% file: sec/0_abstract.tex
\begin{abstract}
Tool-handover interaction calls for preparation before a response is warranted.
The model waits until interaction cues appear, then prepares while assessing whether the accumulating evidence supports a response or calls for cancellation.
We introduce \eb{}, a framework that integrates predictive visual representations into a vision-language interaction model for joint state and tool prediction.
A trainable evidence interface maps features from a frozen V-JEPA~2 encoder into the language decoder alongside native visual inputs, preserving their temporal context and using only observations available at each decision time.
The framework has two instantiations. 
\ebe distinguishes recent from historical evidence and encodes the timing and spatial location of each observation.
\ebi preserves the temporal order of pooled V-JEPA~2 features and interaction turns and generates dedicated interaction tokens.
We evaluate on \bench{}, a streaming-video benchmark for preparation, response, and cancellation in tool-handover assistance.
The best-performing variants of \ebe and \ebi achieve Overall scores of 0.6448 and 0.5892, compared with 0.2178 and 0.5327 for their respective task-adapted native baselines.
These gains span target handovers, long waits, and cancelled requests, demonstrating the value of predictive visual evidence for interaction decisions.
\end{abstract}

%% file: sec/1_intro.tex
\section{Introduction}
\label{sec:intro}

Streaming vision-language models decide both what to say and when to say it as video arrives~\citep{Chen2024VideoLLM,Wang2024MMDuet,Chen2025LiveCC,Yao2026JoyAI}, with benchmarks evaluating these online decisions~\citep{Lin2024StreamingBench,Song2026InteractionBench}.
Tool-handover assistance also requires preparation \emph{before} the worker signals readiness to receive a tool.
For example, an assistant can use visual cues of an upcoming drilling step to prepare a drill, respond to a subsequent handover signal, and cancel preparation if the worker withdraws the request.

The model remains in \wait{} until interaction cues appear, then enters \prepare{} to assess the developing evidence before a response is warranted.
It proceeds to a tool-aware \response{} when the evidence supports the handover, or cancels preparation and returns to \wait{} when the request is withdrawn.
After cancellation, the model continues observing, and new cues can initiate another round of preparation and assessment.
On \bench{}~\citep{cai2026fugibench}, ten off-the-shelf baselines show difficulty making these decisions: some rarely prepare, while others activate during annotated \wait{} intervals (Sec.~\ref{sec:experiments}).

During preparation, the model needs to infer how the observed action is likely to develop, a goal shared with visual anticipation~\citep{Vondrick2016Anticipate,Girdhar2021AVT,Furnari2019RULSTM}.
We use features from the predictively trained V-JEPA~2 encoder to inform these decisions~\citep{Bardes2024VJEPA,Assran2025VJEPA2}.
Later observations can confirm the anticipated handover or indicate that the request has been withdrawn.
Preserving observation timing and order helps the decoder assess which cues remain relevant to the current interaction.

\begin{figure*}[htbp]
\centering
\includegraphics[width=1\textwidth]{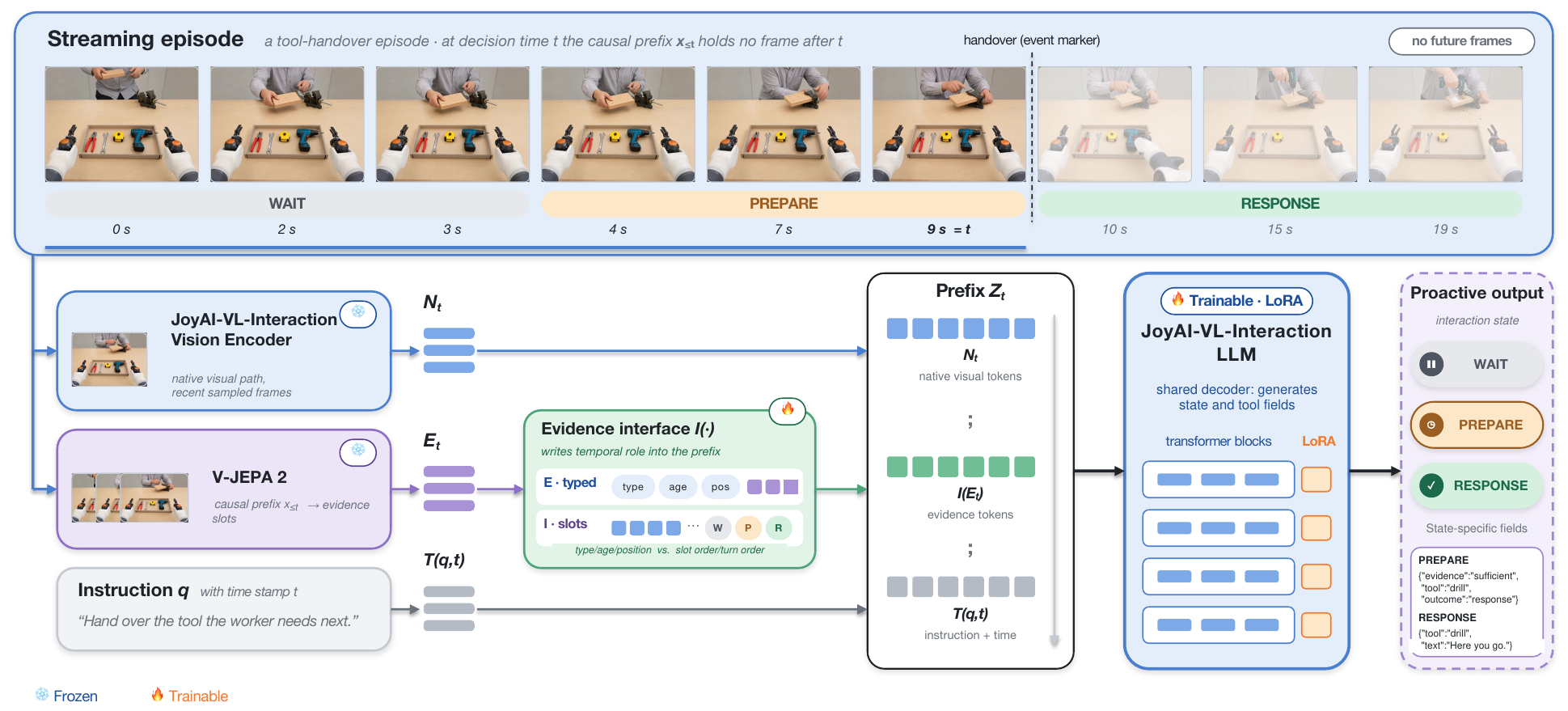}
\caption{\eb{} framework. Top: a tool-handover episode from the test set, illustrating per-second interaction decisions using only frames available at time $t$. Bottom: frozen native and V-JEPA~2 encoders provide visual features to the interaction model. The evidence interface encodes type, age, and position in \ebe, or preserves slot order and turn order in \ebi. The shared decoder, adapted with LoRA, generates interaction states and tool fields.}
\label{fig:architecture}
\end{figure*}

We introduce \eb{} (Fig.~\ref{fig:architecture}), a framework that integrates predictive visual evidence into a vision-language interaction model for joint state and tool generation.
Its trainable evidence interface projects features from a frozen V-JEPA~2 encoder into the language decoder alongside native visual tokens.
We term this \emph{causal visual evidence encoding}: the features use only observations available at the decision time and retain their timing or order in the decoder input.
The framework has two instantiations.
\ebe encodes three attributes: type distinguishes recent from historical evidence, age describes the observation interval relative to the decision time, and position identifies its spatial pooling cell.
\ebi retains temporal context through slot order and turn order, using pooled JEPA features and prior interaction turns, and expresses state decisions with dedicated interaction tokens.

Our contributions are threefold:
\begin{itemize}[leftmargin=*,itemsep=1pt,topsep=2pt]
    \item \emph{A framework for proactive interaction.}
    We introduce \eb{}, which integrates frozen predictive visual representations into a vision-language interaction model, preserving their temporal context for joint state and tool generation.
    \item \emph{Two designs for causal visual evidence encoding.}
    \ebe explicitly encodes type, age, and position. \ebi preserves slot order and turn order and generates dedicated interaction tokens. Both retain the model's autoregressive generation path.
    \item \emph{Empirical validation on \bench{}.}
    The best-performing E and I variants outperform their respective task-adapted native baselines in Overall score and across target handovers, long waits, and cancelled requests. Comparisons and interventions characterize the roles of visual features, evidence interfaces, and temporal metadata in these interaction decisions.
\end{itemize}

%% file: sec/2_related.tex
\section{Related Work}
\label{sec:related}

\paragraph{Visual anticipation and predictive representations.}
Visual anticipation uses observed video to infer upcoming actions or their timing~\citep{Vondrick2016Anticipate,AbuFarha2018When,Furnari2019RULSTM,Girdhar2021AVT}.
Online action detection recognizes ongoing actions using temporal context from the observed video prefix~\citep{Xu2021LSTR,Zhao2022TeSTra,Wang2023MAT}.
Video representation learning provides features for such reasoning: VideoMAE learns through masked video reconstruction~\citep{Tong2022VideoMAE}, while V-JEPA and V-JEPA~2 learn by predicting representations of video content~\citep{Bardes2024VJEPA,Assran2025VJEPA2}.
Such representations allow an interaction model to exploit predictive structure without explicitly generating future frames.
\eb{} integrates frozen V-JEPA~2 features into a generative interaction model, preserving their temporal context to support preparation, response, and cancellation.

\paragraph{Video understanding and streaming interaction.}
Video-language models use visual observations to answer questions and follow instructions.
VideoChat studies instruction-following over video~\citep{Li2023VideoChat}, while LLaVA-OneVision and Qwen-VL extend multimodal understanding to video inputs~\citep{Li2024LLaVAOneVision,An2026LLaVAOneVision2,Wang2024Qwen2VL,Bai2025Qwen3VL}.
Qwen2.5-Omni, Gemma-3n, and the MiniCPM family further broaden the range of multimodal models available for video-based assistance~\citep{Xu2025Qwen25Omni,GemmaTeam2025Gemma3n,Yao2024MiniCPMV}.
MVBench evaluates temporal understanding across diverse video tasks~\citep{Li2024MVBench}.

Streaming interaction additionally requires deciding when to produce an output as new observations arrive.
VideoLLM-online and MMDuet study online video interaction~\citep{Chen2024VideoLLM,Wang2024MMDuet}, while LiveCC supports streaming video commentary~\citep{Chen2025LiveCC}.
MMDuet2, VideoChat3, and MOSS-Video extend this line of streaming interaction~\citep{Wang2026MMDuet2,Li2026VideoChat3,Wang2026MOSSVideo}.
We build on the streaming interaction capabilities of JoyAI-VL-Interaction~\citep{Yao2026JoyAI}.
Relating current observations to past events is another part of streaming understanding, addressed by hierarchical event memory for online temporal grounding~\citep{Zheng2025HEM}.
For tool-handover assistance, the model must also track whether an emerging request remains active as new observations arrive.
\eb{} supplies the interaction decoder with predictive visual evidence and its temporal context to reassess the developing request.

\paragraph{Evaluating proactive interaction.}
StreamingBench evaluates online video comprehension, and InteractionBench examines real-time interaction~\citep{Lin2024StreamingBench,Song2026InteractionBench}.
Evaluating proactive assistance also requires measuring sustained preparation and its withdrawal when a request is cancelled.
\bench{}~\citep{cai2026fugibench} evaluates these behaviors through target handovers, long waits, and cancelled requests.
It measures response-aligned recognition, preparation, and silence over complete interaction trajectories.
We use \bench{} to evaluate how bridged visual evidence supports these interaction decisions.

%% file: sec/3_method.tex
\section{\eb{} Framework}
\label{sec:method}

\subsection{Problem: Causal Future-Grounded Interaction}
\label{sec:problem}

We follow the \emph{\underline{F}uture-\underline{G}rounded \underline{I}nteraction} (\fgi{}) setting of \bench{}~\citep{cai2026fugibench}.
Let $x_\tau$ denote a video frame observed at timestamp $\tau$. At decision time $t$, the model receives a task instruction $q$ and the temporally ordered video prefix $x_{\leq t}$, containing the available frames with $\tau\leq t$.
The output is
\begin{equation}
y_t=(s_t,o_t),\qquad s_t\in\{\wait,\prepare,\response\}.
\end{equation}
Here, $s_t$ is the interaction state and $o_t$ contains its associated structured fields, including tool identity, preparation evidence and outcome, or response text as applicable.
Both \ebe and \ebi serialize $y_t$ as JSON. For \wait{}, the JSON contains only the \texttt{state} field. For \prepare{} and \response{}, it also contains the corresponding fields in $o_t$.

The model remains in \wait{} while no active interaction cue is present.
A cue initiates \prepare{}, during which the model assesses whether the developing evidence is sufficient for a response.
It transitions to a tool-aware \response{} when the handover is due, or returns to \wait{} when the request is withdrawn. After cancellation, it continues observing for new cues that may initiate another cycle.

\subsection{Framework Design}
As shown in Fig.~\ref{fig:architecture}, \eb{} combines native visual features with predictive evidence from a frozen V-JEPA~2 encoder.
The frozen native visual path $F_{\mathrm{native}}$ produces $N_t=F_{\mathrm{native}}(x_{\mathcal W_t})$ from recent frames sampled by time $t$. Here, $\mathcal W_t$ is their timestamp set and $x_{\mathcal W_t}$ their temporally ordered sequence.
The V-JEPA~2 branch~\citep{Assran2025VJEPA2,Bardes2024VJEPA} produces an evidence-token sequence $E_t=\phi(x_{\leq t})$, where $\phi$ denotes clip selection from the available prefix followed by frozen visual encoding.
A trainable interface $I(\cdot)$ maps the JEPA features into the language embedding space, preserving explicit observation attributes in \ebe or slot order and turn order in \ebi.
The current-turn input $Z_t$ combines the two visual streams with the task instruction:
\begin{equation}
Z_t=[N_t;\;I(E_t);\;T(q,t)],
\end{equation}
where semicolons denote concatenation and $T(q,t)$ encodes the instruction $q$ and decision time $t$.
The decoder generates $y_t$ autoregressively through its original language head, using $Z_t$ and any retained interaction context.

Both \ebe and \ebi use the same autoregressive FGI task loss. For the $m$ target tokens $w_1,\ldots,w_m$ encoding $y_t$,
\begin{equation}
\mathcal{L}_{\mathrm{FGI}}=-\sum_{i=1}^{m}\log p_\theta(w_i\mid w_{<i},Z_t),
\end{equation}
where $w_{<i}$ denotes preceding tokens and $p_\theta$ is the conditional token probability under model parameters $\theta$. Conditioning on retained interaction context is implicit.

The native visual encoder and original language-backbone weights remain frozen, as does the V-JEPA~2 encoder when used.
Training updates language LoRA and the evidence interface in variants that include it.
\ebi additionally trains the embeddings and output-head rows associated with its interaction tokens.
In \ebe, LoRA updates the attention query and value projections (\texttt{q\_proj} and \texttt{v\_proj}) of the language decoder. We refer to this configuration as q/v LoRA.

\subsection{\ebe{}: Typed Evidence}
\label{sec:explicit}

In the E-RecentHist configuration, \ebe organizes visual evidence into a recent block and a sequence of historical blocks.
The recent block encodes eight sampled frames as 16 tokens. Historical evidence comprises up to eight complete four-frame chunks, each encoded as four tokens.
The evidence sequence is $E_t^{\mathrm{exp}}=(e_{t,1},\ldots,e_{t,K})$, where $e_{t,k}$ denotes the feature at position $k$ among $K$ token positions.
Only available historical blocks are inserted, in temporal order. During batching, language sequences are padded and padding positions are masked.

The evidence interface projects each 1{,}024-dimensional feature $e_{t,k}$ into the decoder space and adds type, age, and position encodings:
\begin{equation}
b_{t,k}=W_2\sigma(W_1\operatorname{LN}(e_{t,k}))+
u_{\mathrm{type}(k)}+
g_{\mathrm{age}}(a_{t,k})+
u_{\mathrm{pos}(k)},
\end{equation}
where $W_1$ and $W_2$ are learned projection matrices, $\operatorname{LN}$ denotes LayerNorm, and $\sigma$ is GELU. All four terms are 4{,}096-dimensional vectors, added elementwise to form the encoded token $b_{t,k}$.
The learned embedding $u_{\mathrm{type}(k)}$ distinguishes recent from historical tokens.
The position index $\mathrm{pos}(k)$ selects one of four cells in the $2\times2$ spatial pooling grid. Its learned embedding $u_{\mathrm{pos}(k)}$ is reused for the same cell across temporal bins and history blocks.

Age is computed from the observation timestamps rather than selected from an embedding table.
For a token covering the interval from $\tau^{\mathrm{start}}_{t,k}$ to $\tau^{\mathrm{end}}_{t,k}$, its three-dimensional age vector is
\begin{equation}
a_{t,k}=\log\!\left(1+
\begin{bmatrix}
t-\tau^{\mathrm{start}}_{t,k}\\
t-\tau^{\mathrm{end}}_{t,k}\\
\tau^{\mathrm{end}}_{t,k}-\tau^{\mathrm{start}}_{t,k}
\end{bmatrix}\right).
\end{equation}
The components measure elapsed time since the interval started, elapsed time since it ended, and interval duration, with the logarithm applied elementwise.
A learned multilayer perceptron $g_{\mathrm{age}}$ maps $a_{t,k}$ to the decoder dimension.
The encoded tokens form $I_{\mathrm{exp}}(E_t^{\mathrm{exp}})=(b_{t,1},\ldots,b_{t,K})$, giving the decoder input
$Z_t^{\mathrm{exp}}=[N_t;I_{\mathrm{exp}}(E_t^{\mathrm{exp}});T(q,t)]$.

Training proceeds in two stages. Alignment updates only the evidence interface using interaction targets and native-teacher QA responses.
Joint training then optimizes the interface and q/v LoRA with
\begin{equation}
\mathcal{L}_{\mathrm{joint}}=\mathcal{L}_{\mathrm{FGI}}+\lambda_{\mathrm{QA}}\mathcal{L}_{\mathrm{QA}},
\end{equation}
where $\mathcal{L}_{\mathrm{QA}}$ distills QA responses from the native teacher, and $\lambda_{\mathrm{QA}}$ weights this auxiliary loss.

\paragraph{Explicit variants.}
\ebe-RecentHist uses the full interface with recent and historical V-JEPA~2 evidence.
\ebe-Recent retains only the 16 recent tokens, removing all history blocks.
\ebe-Visual replaces V-JEPA~2 features with features from the native visual encoder of JoyAI-VL-Interaction, retaining the same evidence interface, token count, and recent--history layout as \ebe-RecentHist.
\ebe-Native uses only the native visual path and trains q/v LoRA, omitting both the evidence interface and its alignment stage.

\subsection{\ebi{}: Ordered Slots and Interaction Tokens}
\label{sec:implicit}

In the I-MLP configuration, the frozen V-JEPA~2 encoder processes a 16-frame clip ending at the current frame at each interaction turn. Shorter clips are completed by repeating the oldest observed frame.
The encoded features are pooled into eight temporal bins with four spatial cells each, forming the 32-slot sequence $E_t^{\mathrm{imp}}$.
Slots follow temporal order, with spatial cells ordered within each bin.

The evidence interface $I_{\mathrm{imp}}$ normalizes each slot with LayerNorm and maps it into the decoder space using a two-layer MLP with GELU activation.
The current-turn input is
\begin{equation}
Z_t^{\mathrm{imp}}=[N_t;I_{\mathrm{imp}}(E_t^{\mathrm{imp}});T(q,t)].
\end{equation}
The decoder processes this input together with earlier interaction turns retained in chronological order.
Slot order preserves temporal structure within the current clip, while turn order retains the sequence of earlier observations and interaction outputs.

We register \waittok{}, \preparetok{}, and \responsetok{} as special tokens for the three interaction states.
We initialize the embeddings of \waittok{} and \responsetok{} from the native \texttt{</silence>} and \texttt{</response>} embeddings of JoyAI-VL-Interaction, respectively.
The embedding of \preparetok{} is initialized to their arithmetic mean.

\paragraph{Implicit variants.}
\ebi-Hier concatenates V-JEPA~2 features extracted at multiple temporal scales from the observed prefix, then projects them into the JoyAI-VL-Interaction decoder space.
\ebi-Orth replaces the MLP with LayerNorm followed by a linear projection and adds an orthogonality loss to $\mathcal{L}_{\mathrm{FGI}}$ to encourage an approximately orthogonal mapping.
\ebi-Orth+Align retains this architecture and adds a loss aligning the projected V-JEPA~2 features with native visual features from JoyAI-VL-Interaction.
\ebi-Native omits the V-JEPA~2 branch and evidence interface, using only the native visual path while retaining language LoRA and the three interaction tokens.

\begin{figure}[htbp]
\centering
\includegraphics[width=1\linewidth]{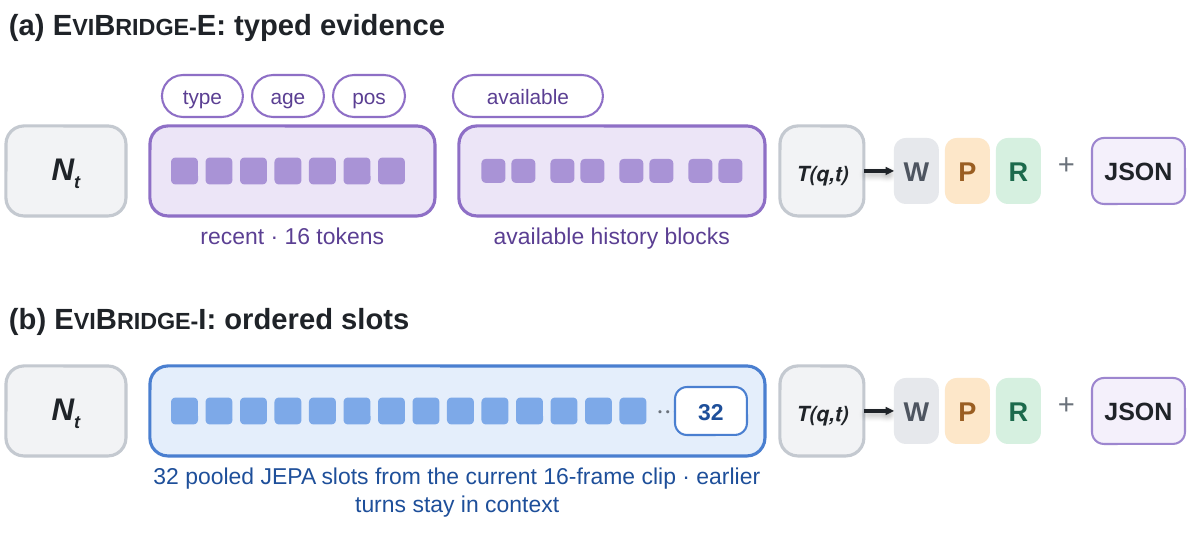}
\caption{Evidence interfaces in \eb{}, shown for (a) E-RecentHist and (b) I-MLP. The explicit interface adds type, age, and position encodings to recent and historical V-JEPA~2 features. The implicit interface retains ordered V-JEPA~2 slots and earlier interaction turns in the decoder context. Both generate JSON with a state field and state-dependent fields. W, P, and R denote \wait{}, \prepare{}, and \response{}, respectively.}
\label{fig:e-vs-i}
\end{figure}

Figure~\ref{fig:e-vs-i} contrasts how \ebe and \ebi organize visual evidence and preserve its temporal context.

%% file: sec/4_experiments.tex
\section{Experiments}
\label{sec:experiments}

We evaluate \eb{} on \bench{}~\citep{cai2026fugibench} against task-adapted native baselines and ten off-the-shelf baselines across target handovers, long waits, and cancelled requests.
Ablations examine evidence configurations, interface designs, and the model's dependence on type, age, and position encodings through inference-time interventions.
We further analyze preparation lead, response timing, cancellation recognition, and false activation to explain the observed performance differences.

\subsection{Experimental Setup}
\paragraph{Benchmark.}
\bench{}~\citep{cai2026fugibench} comprises 19{,}893 synthetic tool-assistance videos covering ten tool categories.
We evaluate on its fixed set of 1{,}000 episodes, comprising 600 target handovers, 200 long waits, and 200 cancelled requests.
Long waits depict ordinary activity without a handover request.
The videos follow human-curated scenario templates, with state boundaries given by the cuts between generated stages. During evaluation-set construction, stage durations are rescaled and the boundary timestamps adjusted to match.
For our experiments, video replay runs at 4\,FPS, while the JoyAI-based models decode once per second from frames available up to that time.

\paragraph{Metrics.}
We report the Overall score on \bench{} along with response-aligned recognition $R$, preparation $P$, and silence $S$.
Each predicted state is held until the next update or the end of the reference stream. All durations are measured in video seconds.
Each episode is labeled as a target handover, long wait, or cancelled request. The benchmark derives the predicted class from the full output trajectory.
For target episode $i$, recognition is $R_i=C_iA_i$, combining event and tool correctness with response alignment.
The indicator $C_i$ is 1 if both the event class and tool are correct, and 0 otherwise. Tool identity is read from the earliest \prepare{} or \response{} observation strictly before handover onset.
The alignment term $A_i$ is the temporal IoU between the union of predicted \response{} intervals and the reference \response{} interval, with $A_i=0$ if no response is predicted.
For long-wait and cancelled episodes, $R_i$ is 1 for a correct event class and 0 otherwise.

Preparation is scored as $P_i=\min(1,\ell_i/(10\,\mathrm{s}))$.
The lead $\ell_i$ measures the time from the start of the final uninterrupted \prepare{} run to handover onset. The run must include the last observation strictly before handover onset. If no such run exists, $\ell_i=0$.
A \wait{}, \response{}, or invalid state interrupts the run.

\begin{figure}[htbp]
\centering
\includegraphics[width=1\linewidth]{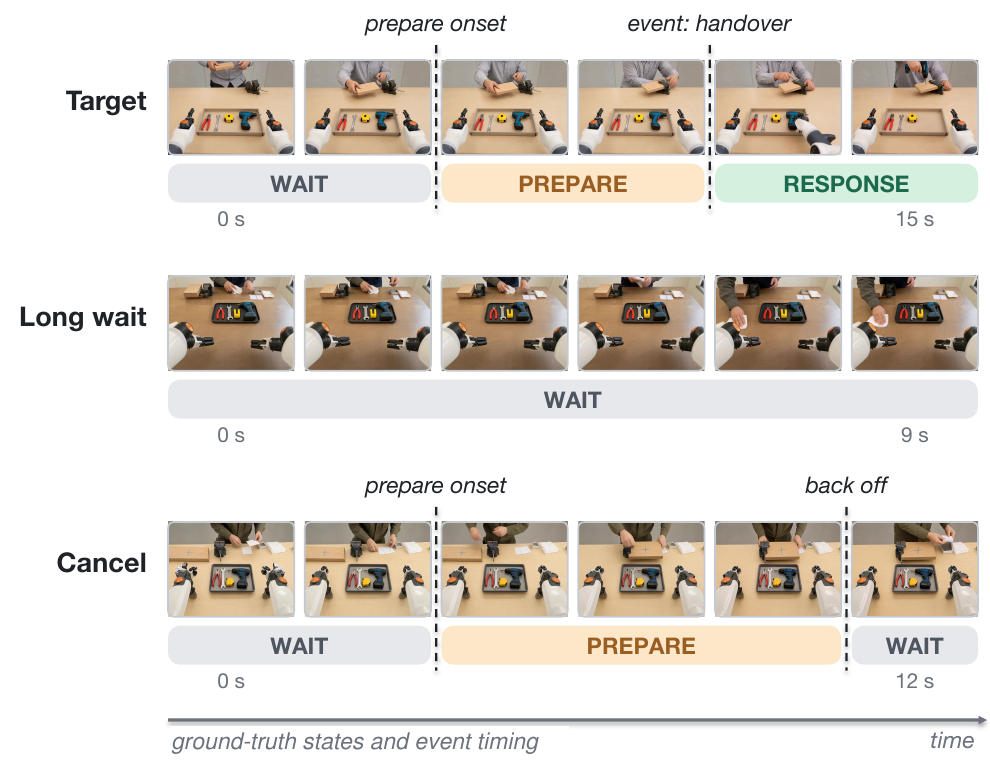}
\caption{Ground-truth state trajectories for three \bench{} test episodes. Top: \prepare{} transitions to \response{} at handover onset. Middle: \wait{} persists throughout ordinary activity. Bottom: cancellation returns the state from \prepare{} to \wait{} without a handover.}
\label{fig:states}
\end{figure}

Silence is scored as $S_i=\max(0,1-f_i/w_i)$, where $w_i$ is the total reference \wait{} duration and $f_i$ is the predicted \prepare{} or \response{} duration within it. We set $S_i=1$ if $w_i=0$.
In cancelled episodes, activity after the reference returns to \wait{} also contributes to $f_i$.
The per-episode score is
\begin{equation}
\mathrm{Overall}_i=\begin{cases}(R_i+H(P_i,S_i))/2,&\text{target},\\(R_i+S_i)/2,&\text{otherwise},\end{cases}
\end{equation}
where $H$ denotes the harmonic mean, with value zero if either argument is zero.
Overall, $R$, and $S$ are averaged over all 1{,}000 episodes. $P$ is averaged over all 600 targets, including zero leads.
Results by episode type report the mean of $\mathrm{Overall}_i$ within each subset.
Figure~\ref{fig:states} illustrates the reference state trajectories for the three episode types using test-set examples.

\paragraph{Training data and splits.}
Training videos follow the same staged generation pipeline and provide state and tool supervision at one-second intervals.
All videos derived from a common source remain in the same split, and evaluation source families are excluded from training.
The E series uses 14{,}002 training and 1{,}721 validation videos.
The I series uses 11{,}093 training and 1{,}232 validation sequences. Each sequence contains the visual inputs and supervised outputs of successive interaction turns from one video.

\paragraph{Shared settings.}
Both series adapt JoyAI-VL-Interaction through language LoRA while keeping its native visual encoder and original language weights frozen. Variants with V-JEPA~2 also freeze that encoder.
LoRA uses rank $r=8$ and scaling $\alpha=16$ in \ebe, and $r=32$ and $\alpha=64$ in \ebi, with dropout 0.05 throughout.
For \ebe, interface alignment precedes joint training of the interface and q/v LoRA. E-Native omits the interface and trains q/v LoRA directly.
\ebi jointly trains language LoRA, the interaction-token rows, and the evidence interface, which is absent in \ebi-Native.

\begin{table*}[htbp]
\caption{Results on \bench{} across 1{,}000 test episodes. $R$, $P$, and $S$ denote response-aligned recognition, preparation, and silence. The final three columns report Overall by episode type. Off-the-shelf baselines are evaluated without task-specific training. Shading highlights the best Overall model in each E/I series. \bestcap{Bold} marks the best value in each column.}
\label{tab:main-baselines}
\centering
\normalsize\setlength{\tabcolsep}{8pt}\renewcommand{\arraystretch}{1.21}
\resizebox{0.98\textwidth}{!}{%
\begin{tabular*}{\textwidth}{@{\extracolsep{\fill}}l*{7}{r}}
\toprule
& & \multicolumn{3}{c}{Components} & \multicolumn{3}{c}{Overall by episode type}\\
\cmidrule(lr){3-5}\cmidrule(lr){6-8}
Method & Overall$\uparrow$ & $R\uparrow$ & $P\uparrow$ & $S\uparrow$ & Target$\uparrow$ & Long wait$\uparrow$ & Cancel$\uparrow$\\
\midrule
\multicolumn{8}{c}{\textit{Off-the-shelf baselines}}\\
Qwen2.5-Omni-7B~\cite{Xu2025Qwen25Omni} & 0.3012 & 0.2046 & 0.0982 & 0.8948 & 0.0144 & 0.9769 & 0.4857\\
JoyAI-VL-Interaction~\cite{Yao2026JoyAI} & 0.3002 & 0.2003 & 0.0000 & 0.9990 & 0.0003 & \best{1.0000} & 0.5000\\
LLaVA-OneVision-2-8B~\cite{An2026LLaVAOneVision2} & 0.2996 & 0.2002 & 0.0046 & 0.9954 & 0.0002 & \best{1.0000} & 0.4975\\
MMDuet2~\cite{Wang2026MMDuet2} & 0.2995 & 0.2000 & 0.0000 & 0.9989 & 0.0000 & \best{1.0000} & 0.4974\\
Qwen3-VL-8B-Instruct~\cite{Bai2025Qwen3VL} & 0.2979 & 0.2126 & 0.0044 & 0.9161 & 0.0410 & 0.8995 & 0.4668\\
LiveCC-7B-Instruct~\cite{Chen2025LiveCC} & 0.2173 & 0.0365 & 0.0000 & 0.9978 & 0.0012 & 0.5859 & 0.4969\\
VideoChat3-4B~\cite{Li2026VideoChat3} & 0.1999 & 0.0000 & 0.0000 & \best{0.9998} & 0.0000 & 0.5000 & 0.4994\\
MOSS-Video~\cite{Wang2026MOSSVideo} & 0.1276 & 0.0232 & 0.0509 & 0.6774 & 0.0407 & 0.2624 & 0.2534\\
Gemma-3n-E4B-it~\cite{GemmaTeam2025Gemma3n} & 0.0920 & 0.0527 & 0.1643 & 0.2196 & 0.0419 & 0.1406 & 0.1938\\
MiniCPM-o-2.6~\cite{Yao2024MiniCPMV} & 0.0906 & 0.0125 & 0.1073 & 0.3683 & 0.0467 & 0.1166 & 0.1963\\
\midrule
\multicolumn{8}{c}{\textit{EviBridge-E}}\\
E-Native & 0.2178 & 0.0451 & \best{0.3497} & 0.6611 & 0.1790 & 0.3404 & 0.2115\\
E-Visual & 0.5183 & 0.5395 & 0.1497 & 0.8898 & 0.3005 & 0.8213 & 0.8687\\
E-Recent & 0.6253 & 0.7223 & 0.1948 & 0.9200 & 0.4101 & 0.9946 & 0.9017\\
\rowcolor{evirow}E-RecentHist & \best{0.6448} & \best{0.7406} & 0.2077 & 0.8952 & \best{0.4450} & 0.9835 & \best{0.9056}\\
\midrule
\multicolumn{8}{c}{\textit{EviBridge-I}}\\
I-Native & 0.5327 & 0.5873 & 0.1407 & 0.8454 & 0.3735 & 0.9272 & 0.6157\\
I-Hier & 0.5215 & 0.5533 & 0.1391 & 0.8739 & 0.3577 & 0.9529 & 0.5815\\
I-Orth & 0.5609 & 0.6532 & 0.1432 & 0.8232 & 0.3890 & 0.8983 & 0.7392\\
I-Orth+Align & 0.5543 & 0.5658 & 0.2679 & 0.8409 & 0.3551 & 0.9546 & 0.7518\\
\rowcolor{evirow}I-MLP & 0.5892 & 0.6934 & 0.1383 & 0.8537 & 0.3829 & 0.9660 & 0.8311\\
\bottomrule
\end{tabular*}%
}
\end{table*}

\paragraph{Models.}
We use E- and I- to denote the \ebe and \ebi variants described in Secs.~\ref{sec:explicit} and~\ref{sec:implicit}.
Our ten off-the-shelf baselines span general-purpose multimodal understanding and native streaming interaction.
The former group comprises Qwen3-VL~\citep{Bai2025Qwen3VL}, LLaVA-OneVision-2~\citep{An2026LLaVAOneVision2}, Qwen2.5-Omni~\citep{Xu2025Qwen25Omni}, Gemma-3n~\citep{GemmaTeam2025Gemma3n}, and MiniCPM-o-2.6~\citep{Yao2024MiniCPMV}, evaluated with state-and-tool instructions.
The latter comprises JoyAI-VL-Interaction~\citep{Yao2026JoyAI}, MMDuet2~\citep{Wang2026MMDuet2}, LiveCC~\citep{Chen2025LiveCC}, VideoChat3~\citep{Li2026VideoChat3}, and MOSS-Video~\citep{Wang2026MOSSVideo}, retaining their native emission and no-update mechanisms.
All models are evaluated on the same episodes under the \bench{} scoring protocol.

\subsection{Main Results}
As shown in Table~\ref{tab:main-baselines}, the best-performing E and I variants both outperform their respective task-adapted native baselines in Overall.
E-RecentHist achieves 0.6448 compared with 0.2178 for E-Native, while I-MLP achieves 0.5892 compared with 0.5327 for I-Native.

E-RecentHist improves response-aligned recognition and silence over E-Native, whose higher preparation score is accompanied by frequent activation during reference \wait{} intervals.
I-MLP also improves recognition over I-Native while maintaining a similar preparation score.

Qwen2.5-Omni is the strongest off-the-shelf baseline in Overall at 0.3012. Both E-RecentHist and I-MLP outperform it in response-aligned recognition and sustained preparation.
Off-the-shelf baselines vary widely in preparation and false activation.
Several achieve high silence scores but little preparation, whereas Gemma-3n, MiniCPM-o, and MOSS-Video prepare more but are frequently active during reference \wait{}.
The strongest E and I variants achieve higher Overall by combining preparation with more accurate recognition, rather than maximizing preparation or silence alone.

\begin{figure*}[ht]
\centering
\includegraphics[width=\textwidth]{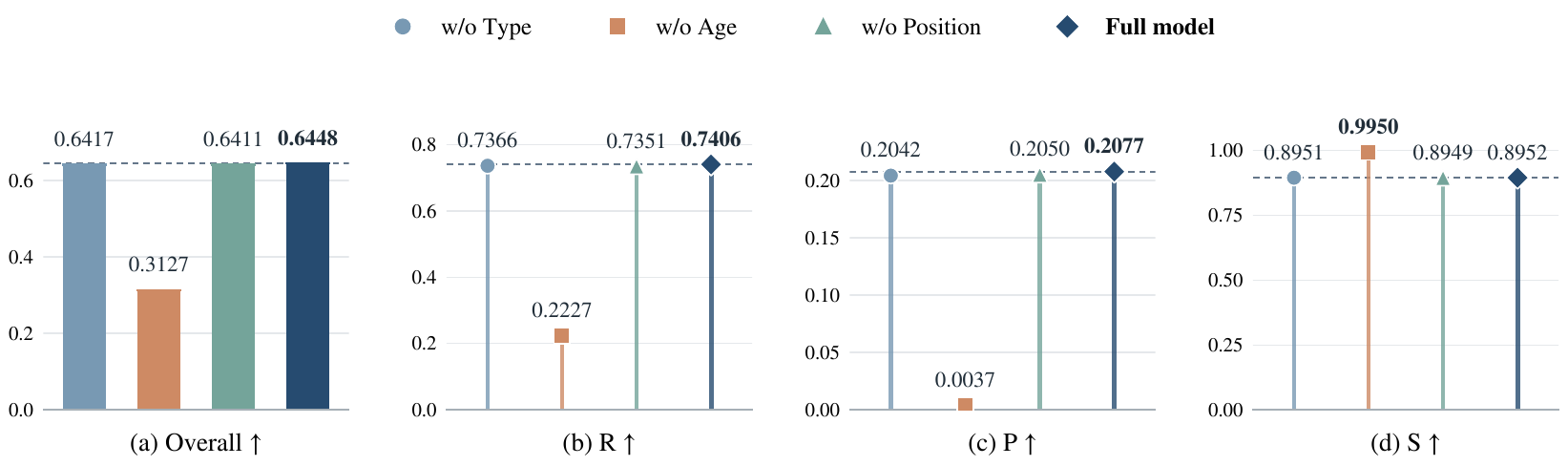}
\caption{Fixed-weight inference interventions on E-RecentHist metadata branches. Each intervention zeros the indicated trained branch for all recent and historical evidence tokens. All axes start at zero. Dashed lines mark the historical full-model scores; bold marks the best score in each panel.}
\label{fig:e-component-ablation}
\end{figure*}

\subsection{Episode-type Analysis}
Table~\ref{tab:main-baselines} shows that E-RecentHist and I-MLP outperform their respective native baselines across all three episode types.
On target handovers, E-RecentHist achieves an Overall score of 0.4450, compared with 0.1790 for E-Native. I-MLP reaches 0.3829, while none of the off-the-shelf baselines exceeds 0.0467.
On long waits, E-Recent and E-RecentHist achieve near-perfect scores, combining accurate no-event classification with sustained \wait{}.
Several off-the-shelf baselines also score perfectly on long waits, but perform poorly on target handovers. The JEPA-equipped E variants handle handovers while retaining strong performance when no request occurs.

I-MLP shows its largest improvement over I-Native on cancelled requests, raising Overall from 0.6157 to 0.8311.
E-Recent and E-RecentHist both exceed 0.90, compared with 0.2115 for E-Native.
Off-the-shelf baselines score at most 0.5000 on cancelled requests, so the advantage of both E-RecentHist and I-MLP extends to episodes in which the anticipated handover does not occur.
These results reflect cancellation recognition and inactivity during reference \wait{} intervals, including those following withdrawal.

\subsection{Ablations}
\paragraph{Evidence configurations and interface designs.}
Table~\ref{tab:main-baselines} shows that E-Visual improves on E-Native by supplying native visual features through an additional evidence interface.
Compared with E-Visual, E-Recent improves recognition, preparation, and silence using only recent V-JEPA~2 features, raising Overall from 0.5183 to 0.6253.
This gain is achieved without historical blocks, supporting the value of recent predictive features.
E-RecentHist achieves higher response-aligned recognition and preparation than E-Recent, yielding a higher Overall despite a lower silence score.

In the I series, the MLP interface achieves higher recognition and Overall than either multi-scale feature concatenation or the orthogonality-regularized linear projection.
Adding visual alignment to I-Orth improves preparation and silence, while recognition and Overall decrease.

\begin{table*}[ht]
\caption{Behavior diagnostics. Row shading follows Table~\ref{tab:main-baselines}.
Acc., F1, and Cancel denote event-class accuracy, macro-F1, and cancellation recall.
Cov. measures the fraction of targets with positive preparation lead. Lead averages $\ell_i$ over all targets, including zeros, without capping.
WAIT-FA measures predicted \prepare{} or \response{} time as a fraction of pooled reference \wait{} time.
Recall measures the fraction of targets with predicted--reference \response{} overlap.
Onset is the mean signed first-response timing error in seconds, evaluated on targets with a predicted \response{}. Positive values indicate delay. \bestcap{Bold} marks the best values except for Onset.}
\label{tab:behavior}
\centering
\normalsize\setlength{\tabcolsep}{5pt}\renewcommand{\arraystretch}{1.21}
\resizebox{0.98\textwidth}{!}{%
\begin{tabular*}{\textwidth}{@{\extracolsep{\fill}}l*{8}{r}}
\toprule
& \multicolumn{3}{c}{Event classification} & \multicolumn{2}{c}{Preparation} & \multicolumn{1}{c}{\shortstack{False\\activation}} & \multicolumn{2}{c}{Response}\\
\cmidrule(lr){2-4}\cmidrule(lr){5-6}\cmidrule(lr){7-7}\cmidrule(lr){8-9}
Method & Acc.$\uparrow$ & F1$\uparrow$ & Cancel$\uparrow$ & Cov.$\uparrow$ & Lead (s)$\uparrow$ & WAIT-FA$\downarrow$ & Recall$\uparrow$ & Onset (s)\\
\midrule
\multicolumn{9}{c}{\textit{Off-the-shelf baselines}}\\
Qwen2.5-Omni-7B~\cite{Xu2025Qwen25Omni} & 0.7600 & 0.5346 & 0.0000 & 0.1233 & 1.187 & 0.0692 & 0.9233 & \phantom{$-$}1.055\\
JoyAI-VL-Interaction~\cite{Yao2026JoyAI} & 0.5660 & 0.4125 & 0.0000 & 0.0000 & 0.000 & \best{0.0003} & 0.6017 & \phantom{$-$}3.079\\
LLaVA-OneVision-2-8B~\cite{An2026LLaVAOneVision2} & 0.2260 & 0.1414 & 0.0000 & 0.0050 & 0.054 & 0.0048 & 0.0400 & \phantom{$-$}1.104\\
MMDuet2~\cite{Wang2026MMDuet2} & 0.3890 & 0.2913 & 0.0000 & 0.0000 & 0.000 & 0.0017 & 0.2483 & \phantom{$-$}4.095\\
Qwen3-VL-8B-Instruct~\cite{Bai2025Qwen3VL} & 0.7680 & 0.5346 & 0.0000 & 0.0600 & 0.044 & 0.0714 & 0.9833 & $-1.722$\\
LiveCC-7B-Instruct~\cite{Chen2025LiveCC} & 0.0450 & 0.0954 & 0.0000 & 0.0000 & 0.000 & 0.0037 & 0.0150 & $-3.513$\\
VideoChat3-4B~\cite{Li2026VideoChat3} & 0.1020 & 0.0967 & 0.0000 & 0.0000 & 0.000 & 0.0004 & 0.1000 & \phantom{$-$}5.167\\
MOSS-Video~\cite{Wang2026MOSSVideo} & 0.3700 & 0.2086 & 0.0050 & 0.2000 & 0.525 & 0.4226 & 0.4900 & $-4.654$\\
Gemma-3n-E4B-it~\cite{GemmaTeam2025Gemma3n} & 0.6030 & 0.2698 & 0.0050 & 0.3400 & 1.894 & 0.7247 & 0.9217 & $-6.007$\\
MiniCPM-o-2.6~\cite{Yao2024MiniCPMV} & 0.6000 & 0.2500 & 0.0000 & 0.2467 & 1.156 & 0.6585 & 0.9917 & $-6.075$\\
\midrule
\multicolumn{9}{c}{\textit{EviBridge-E}}\\
E-Native & 0.6140 & 0.3331 & 0.0050 & \best{0.6767} & \best{3.666} & 0.4355 & 0.4033 & \phantom{$-$}6.165\\
E-Visual & 0.8820 & 0.8148 & 0.8000 & 0.4667 & 1.497 & 0.1309 & 0.8700 & $-0.207$\\
E-Recent & \best{0.9780} & \best{0.9704} & \best{0.9000} & 0.5733 & 1.950 & 0.0992 & \best{1.0000} & \phantom{$-$}0.185\\
\rowcolor{evirow}E-RecentHist & 0.9670 & 0.9482 & 0.8650 & 0.5650 & 2.077 & 0.1206 & \best{1.0000} & \phantom{$-$}0.220\\
\midrule
\multicolumn{9}{c}{\textit{EviBridge-I}}\\
I-Native & 0.8280 & 0.7225 & 0.3550 & 0.4117 & 1.407 & 0.1829 & 0.9317 & $-0.393$\\
I-Hier & 0.7730 & 0.6476 & 0.2400 & 0.3967 & 1.392 & 0.1432 & 0.8867 & $-0.606$\\
I-Orth & 0.8930 & 0.8571 & 0.6650 & 0.4183 & 1.432 & 0.2091 & 0.9850 & $-0.887$\\
I-Orth+Align & 0.9120 & 0.8800 & 0.6900 & 0.5783 & 2.730 & 0.1875 & 0.6950 & \phantom{$-$}0.812\\
\rowcolor{evirow}I-MLP & 0.9330 & 0.9055 & 0.7850 & 0.4200 & 1.383 & 0.1702 & 0.9700 & $-0.729$\\
\bottomrule
\end{tabular*}%
}
\end{table*}

\paragraph{Type, age, and position.}
Figure~\ref{fig:e-component-ablation} examines sensitivity to each attribute by zeroing its branch output at inference for all recent and historical tokens in E-RecentHist. Model weights, visual features, and masks remain unchanged.
Zeroing age reduces Overall to 0.3127 from the historical full-model score of 0.6448 and preparation to 0.0037.
Recognition also falls sharply, while silence rises to 0.9950. The increase in silence therefore does not indicate an improvement in overall interaction performance.
Zeroing type or position produces much smaller changes, reducing Overall by less than 0.004.
Among the three interventions, zeroing age causes the largest decline in preparation and recognition.
The no-age run contains 10 schema-invalid outputs among 19{,}109 model calls.

\subsection{Behavior Diagnostics}
\label{sec:diagnostics}

Table~\ref{tab:behavior} separates episode classification from preparation and response timing.
High response recall can coexist with poor response timing.
Gemma-3n and MiniCPM-o achieve recall above 0.92, but their mean onset errors are about $-6$\,s and their WAIT-FA rates exceed 0.65.
Qwen2.5-Omni has much lower WAIT-FA at 0.0692, but prepares on only 12.33\% of target episodes and has a mean response delay of 1.055\,s.

E-Recent and E-RecentHist prepare on more than half of the target episodes, with mean leads of about two seconds across all targets.
Both achieve response recall of 1.0 and mean signed onset errors of about $+0.2$\,s.
E-Native has a longer mean lead, but its response recall is 0.4033 and WAIT-FA is 0.4355. The corresponding WAIT-FA rates are 0.0992 for E-Recent and 0.1206 for E-RecentHist.
These variants cover more target responses while reducing activation during reference \wait{}.
E-Recent leads in event classification and cancellation recall, whereas E-RecentHist achieves higher response-aligned recognition.
These rankings differ because response-aligned recognition also accounts for tool correctness and temporal overlap with the reference response on target episodes.

I-MLP retains preparation coverage and mean lead close to I-Native. Response recall rises from 0.9317 to 0.9700, and WAIT-FA falls from 0.1829 to 0.1702.
Cancellation recognition shows a larger change, with recall rising from 0.3550 to 0.7850. None of the off-the-shelf baselines exceeds 0.0050 on this measure.

%% file: sec/5_conclusion.tex
\section{Conclusion}
\label{sec:conclusion}

\looseness=-1
We presented \eb{}, a framework that integrates predictive visual evidence into a vision-language interaction model for tool-handover assistance.
Its trainable evidence interface connects frozen V-JEPA~2 features to the native decoder while preserving their temporal context.
\ebe encodes type, age, and position explicitly, while \ebi preserves slot order and turn order and uses dedicated interaction tokens.
Both use the observed video to jointly predict the interaction state and the relevant tool.

On \bench{}, the strongest E and I variants outperform their respective task-adapted native baselines in Overall across all three episode types.
Behavior diagnostics show higher response recall, better cancellation recognition, and less false activation.
Inference-time interventions reveal that E-RecentHist depends more on age than on type or position for preparation and recognition.
These results support integrating predictive visual evidence and its temporal context into interaction decisions.

Future work will explore architectures that adaptively select and aggregate predictive visual evidence for interaction. We will also evaluate these designs across a broader range of scenarios and tasks.